\documentclass{bvm} 

\usepackage{booktabs}
\usepackage{array}
\usepackage{svg}
\usetikzlibrary{positioning, arrows.meta, calc}
\usepackage{tikz}
\usetikzlibrary{positioning, shadows, calc}
\usepackage[hidelinks]{hyperref}

\begin{document}

\newcommand{\bvmyear}{2026}

\selectlanguage{english} 

\title{Benchmarking CNN-based Models against Transformer-based Models for Abdominal Multi-Organ Segmentation on the RATIC Dataset}

\titlerunning{BVM \bvmyear}

\author{
	\fname{Lukas} \lname[]{Bayer} \inst{1} \affiliation{Friedrich-Alexander-Universität Erlangen-Nürnberg} \authorsEmail{lukas.bayer@fau.de} \isResponsibleAuthor,
	\fname{Sheethal} \lname[]{Bhat} \inst{1} \affiliation{Friedrich-Alexander-Universität Erlangen-Nürnberg} \authorsEmail{sheethal.bhat@fau.de}, 
	\fname{Andreas} \lname[]{Maier} \inst{1} \affiliation{Friedrich-Alexander-Universität Erlangen-Nürnberg} \authorsEmail{andreas.maier@fau.de}
	}

\authorrunning{Bayer}

\institute{
	\inst{1} Pattern Recognition Lab Friedrich-Alexander-Universität Erlangen-Nürnberg\\
}

\email{lukas.bayer@fau.de}

\maketitle

\begin{abstract}
	Accurate multi-organ segmentation in abdominal CT scans is essential for computer-aided diagnosis and treatment. While convolutional neural networks (CNNs) have long been the standard approach in medical image 
	segmentation, transformer-based architectures have recently gained attention due to their ability to model long-range 
	dependencies. In this study, we systematically benchmark the three hybrid transformer-based models UNETR, SwinUNETR, and 
	UNETR++ against a strong CNN baseline, SegResNet, for volumetric multi-organ segmentation on the heterogeneous RATIC 
	dataset. The dataset comprises 206 annotated CT scans from 23 institutions worldwide, covering five abdominal organs. 
	All models were trained and evaluated under identical preprocessing and training conditions using the Dice Similarity 
	Coefficient (DSC) as the primary metric. The results show that the CNN-based SegResNet achieves the highest overall 
	performance, outperforming all hybrid transformer-based models across all organs. Among the transformer-based 
	approaches, UNETR++ delivers the most competitive results, while UNETR demonstrates notably faster convergence 
	with fewer training iterations. These findings suggest that, for small- to medium-sized heterogeneous datasets, 
	well-optimized CNN architectures remain highly competitive and may outperform hybrid transformer-based designs.\\
	Code: \url{https://github.com/lukas-FAU/medical_image_segmentation_benchmarking_paper}
\end{abstract}

\section{Introduction}
Abdominal multi-organ segmentation is a cornerstone of modern digital medicine, transforming static pixels from CT scans 
into dynamic, 3D anatomical maps.
\par Over the past decade, convolutional neural networks (CNNs), in particular U-Net–based \cite{0000-03} architectures, have 
become the de facto standard for medical image 
segmentation due to their strong inductive bias toward local spatial context and their data efficiency.

\par More recently, transformer-based \cite{0000-16} models have been introduced to overcome the limited receptive field of 
convolutions by modeling long-range dependencies through self-attention. Hybrid approaches such as UNETR \cite{0000-01}, 
SwinUNETR \cite{0000-04}, and UNETR++ \cite{0000-06} combine transformer encoders with convolutional decoders and have 
reported competitive or superior results on abdominal multi-organ segmentation tasks. 
\par However, benchmarking between CNN-based and transformer-based models is frequently confounded by the practice 
of evaluating baselines while simultaneously proposing a new architecture. Therefore reported performance metrics for 
a single architecture that is used as baseline often exhibits a significant 
inter-publication variance. For instance, the UNETR \cite{0000-01} model which is commonly used as a baseline  yields 
reported DSC scores ranging from 0.6888 \cite{0000-22} to 0.8910 \cite{0000-20, 0000-01}, with intermediate results of 
0.7835 \cite{0000-17}, 0.8027 \cite{0000-18}, and 0.8129 \cite{0000-19} while trained on the BTCV dataset \cite{0000-21} with 
comparable hyperparameters.
This variance makes it difficult to draw conclusions about the actual performance of architectures on abdominal multi-organ 
segmentation tasks. This also complicates a 
systematic comparison of the performance of CNN-based and transformer-based models. 
Therefore independent performance benchmarks are important to provide a fair and comprehensive comparison of different 
architectures under identical conditions.
\par While several meta-analyses \cite{0000-25, 0000-26} evaluating the performance of transformer-based models compared 
to CNN-based models exist,
the field lacks a comprehensive comparison with independent benchmarks. To the best of our knowledge, currently only one 
independent benchmark has been published comparing the segmentation performance of CNN-based and transformer-based models on
abdominal CT datasets. 
This study \cite{0000-23} by Isensee et al. has shown that CNN-based models systematically outperform transformer-based models in abdominal 
multi-organ segmentation tasks when trained on
small datasets of 50 scans while certain architectures achieve almost the same accuracy when trained on larger datasets with
over 500 scans. However, while including 12 CNN-based models it only benchmarks 4 transformer-based models, leaving out some
transformer-based architectures like UNETR \cite{0000-01} and UNETR++ \cite{0000-06} that reported competitive performance 
in their original publications. Furthermore, it did not include medium-sized datasets of around 200 scans. Since it is often
claimed that the size of the training dataset is a crucial factor for the performance of transformer-based models \cite{0000-16}, 
it is important to evaluate the performance of these models on medium-sized datasets as well. 

\par Therefore in this work, we systematically benchmark the three hybrid transformer-based architectures UNETR, SwinUNETR and 
UNETR++ against a strong CNN baseline, 
SegResNet, for abdominal multi-organ segmentation on the medium-sized and heterogeneous RATIC dataset. By evaluating segmentation accuracy, convergence 
behavior across heterogeneous CT scans, we aim to provide insights into whether transformer-based designs 
offer tangible advantages over well-optimized CNNs in this setting.

\section{Materials and Methods}
\subsection{Dataset details}
For this benchmark we used the RATIC dataset \cite{0000-08}, which was published by the Radiological Society of North America (RSNA) in 2023.
The dataset provides in total 206 pixel-level segmentation labels of the liver, spleen, left kidney,
right kidney, and bowel in Neuroimaging Informatics Technology Initiative (NIfTI) format. The labels were created by a nnU-Net model \cite{0000-09}
trained on the TotalSegmentator dataset \cite{0000-10} and manually corrected by medical experts. The CT images are provided in Digital Imaging and
Communications in Medicine (DICOM) format. Each scan has a size of 512x512 pixels in the axial plane, while the number of slices in the z-direction varies between 50 and 300.
\par
\begin{figure}[H]
	\centering
	\includegraphics[width=0.8\textwidth]{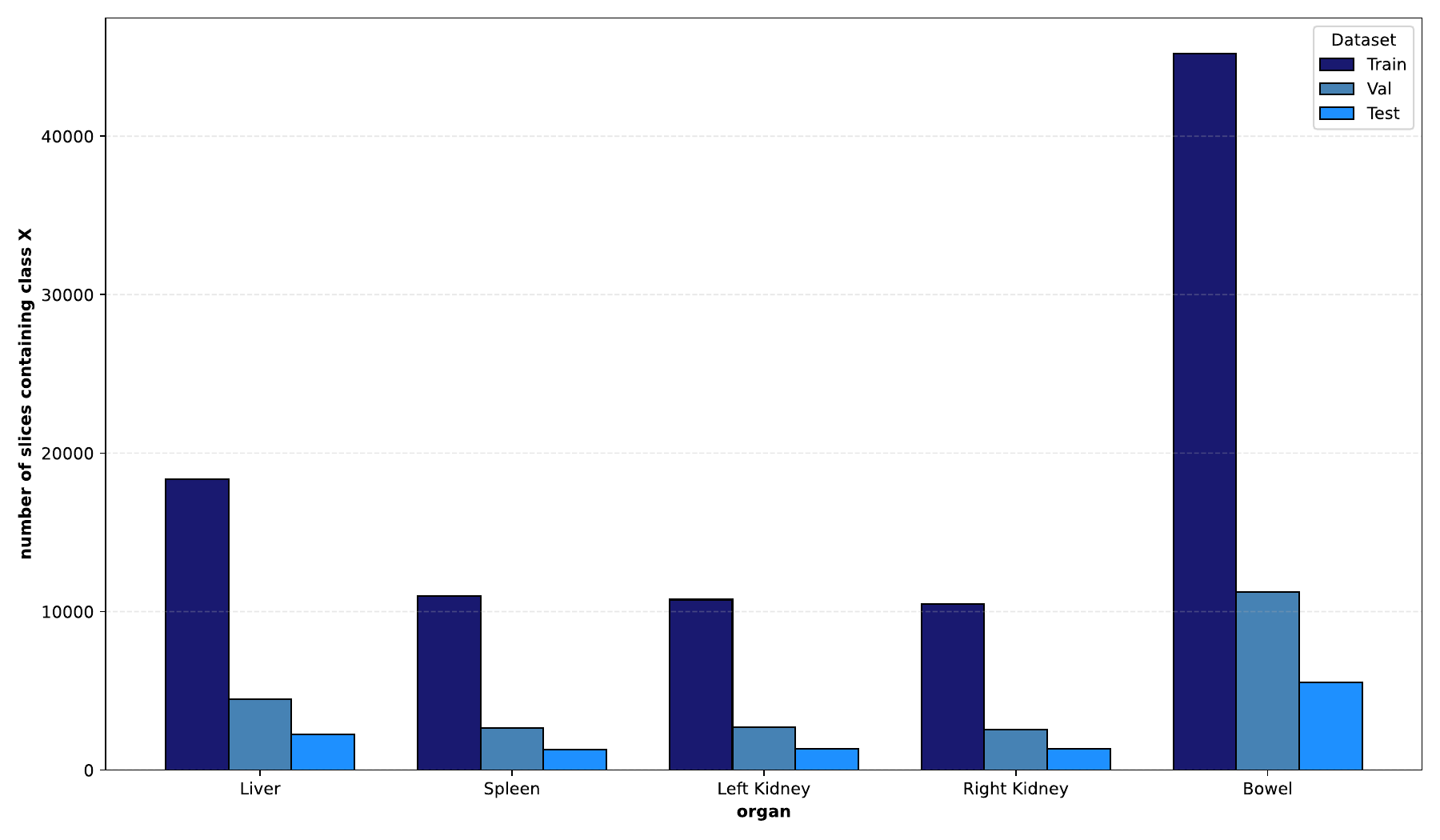}
	\caption{Class wise distribution for train, test and validation dataset}
	\label{Fig-1}
\end{figure}

 We split the dataset into a training set of 144 scans, a validation set of 41 scans and a test set of 21 scans. Figure \ref{Fig-1} shows the 
class-wise distribution of the training, validation and test set. The dataset is imbalanced, with larger organs like the bowel and the liver
appearing more frequently than smaller organs like the kidneys or the spleen.

Before training, the CT scans and labels were preprocessed. The following steps were performed:
\paragraph{\textbf{Orientation}}
The NIfTI files for the labels use a Right-Anterior-Superior (RAS) coordinate system. However, the CT images, which are in DICOM format,
use a Left-Posterior-Superior (LPS) coordinate system. Therefore, the CT images were reoriented to match the RAS system of the labels.
\paragraph{\textbf{Spacing}}
The RATIC dataset \cite{0000-08} contains CT scans from 23 institutions across 14 countries
and six continents. Therefore, it can be assumed that the scans were acquired using different
CT scanners with varying acquisition protocols. As a result, the voxel spacing can vary significantly between scans.
To ensure that all images have the same voxel spacing, we resampled all images and
segmentations to an isotropic spacing of $1 mm \times 1 mm \times 1 mm$.
\paragraph{\textbf{Normalization}}
While the NIfTI files already have an intensity scale in the range from 0 to 1, as
it is needed for training neural networks, the DICOM files in the RATIC dataset
stores the intensity values in a range from -150 to 250. Therefore, we normalized
the intensity values of the CT images to the range $[0, 1]$.
\paragraph{\textbf{Class Imbalance}}
For training we did not use the whole 3D volume of the CT scans, but randomly sampled patches of size $96 \times 96 \times 96$ voxels.
However, due to the varying sizes of the organs and the high amount of background voxels, there is a significant class imbalance.
To address this, we used a patch sampling strategy that ensures that patches containing smaller organs are sampled more frequently.
We used the following sampling ratios: [1:1:2:2:2:1] for [background, liver, left kidney, right kidney, spleen, bowel]. This means
that patches containing the kidneys or spleen are sampled twice as often as patches containing only background, liver or bowel.
\par We also tried a similar ratio for the weighting of the loss function during training, but this made the training less stable
while not leading to an improvement in performance.
\par A different approach we tested was to increase the size of the patches to $128 \times 128 \times 128$ voxels. This way, each patch
contains more context information and the organs occupy a larger portion of the patch. However, due to the limited GPU memory, this only
allowed for a batch size of 1 during training, which made the training less stable and did not lead to an improvement in performance.
\paragraph{\textbf{Generalization}}
To improve the generalization of the models, we used random flips along all three axes as well as random rotations of 90, 180 and 270 degrees during training.
We also applied random intensity scaling and shifting to the input images.

\subsection{The Segmentation Models}
This assessment evaluates four state-of-the-art models for medical image segmentation, comprising one CNN-based model
and three hybrid models which use a transformer-based encoder combined with a CNN-based decoder.
\paragraph{\textbf{UNETR}} UNETR \cite{0000-01} is a U-Net \cite{0000-03} style segmentation model. It was one of the first architectures
which successfully completely replaced the encoder of a U-Net with a Vision Transformer (ViT) \cite{0000-02}. In the encoder,
the 3D input volume is split into non-overlapping patches, which are then flattened and projected into a 1D sequence of embeddings.
Then positional embeddings are added to the tokens and the sequence is passed through multiple transformer layers. Through skip connections
feature maps are extracted from layers 3, 6, 9 and 12 of the transformer encoder. The decoder remains convolutional. It is
a hierarchical CNN that upsamples the feature map extracted by the transformer encoder's last layer and combines it with the
feature maps that were extracted from different encoder stages via skip connections. Finally, when the decoder reaches the
original input resolution, a 1x1x1 convolution is used to project the features into the desired number of output classes.
\par Since it uses transformer blocks in the encoder, UNETR is able to capture long-range dependencies and global context information.
However, it also has a large number of parameters and requires a larger amount of training data to achieve good performance.
\paragraph{\textbf{SwinUNETR}} SwinUNETR \cite{0000-04} is also a U-Net style architecture. Similar to UNETR, it replaces the encoder
of a U-Net with a transformer-based architecture. However, instead of using a ViT as encoder, it uses a Swin Transformer \cite{0000-05}.
The Swin Transformer uses shifted windows to compute self-attention locally, which reduces the computational complexity
compared to global self-attention. The Swin Transformer encoder consists of four stages, each reducing the resolution
of the input by a factor of two. Feature maps are extracted from each of the four stages and passed to the convolutional decoder
via skip connections. The decoder is similar to the one used in UNETR. It progressively upsamples the feature maps and combines them with the
corresponding encoder features. A final 1x1x1 convolution is used to produce the segmentation output.
\par The Shifted window attention reduces complexity from quadratic to linear relative to image size, promising an increase in training speed.
\paragraph{\textbf{UNETR++}} UNETR++ \cite{0000-06} is an improved version of UNETR. It retains the overall architecture of UNETR,
but uses several enhancements to improve performance. While UNETR uses pure vision transformer blocks in the encoder,
UNETR++ uses a hybrid approach that combines convolutional layers with transformer blocks. First the input is downsampled before
passing through an efficient paired-attention (EPA) block. The EPA block computes two types of attention: spatial attention
and channel attention, which are then combined to produce the output feature map. This allows the model
to capture both local and global features more effectively. Additionally, UNETR++ uses a decoder that is symmetric to the encoder,
with skip connections between corresponding encoder and decoder layers. This helps to preserve spatial information and improve segmentation
accuracy. Like UNETR, a final 1x1x1 convolution is used to produce the segmentation output.
\par UNETR++ is designed to improve efficiency and accuracy compared to UNETR. However, it also has a larger number of parameters and may require 
more training data to achieve good performance.
\paragraph{\textbf{SegResNet}} SegResNet \cite{0000-07} is a CNN-based architecture that extends the U-Net design by adding
residual blocks. Unlike standard symmetric U-Net, the encoder is significantly larger and deeper than the decoder.
It uses ResNet-style blocks where each block consists of Group Normalization, a ReLU activation, 3D convolutions and a skip connection.
The decoder uses 3D bilinear upsampling followed by a 1x1x1 convolutions to reduce the number of feature channels before concatenating them
with skip connections from the encoder. The SegResNet architecture proposed in the original paper also uses a variational
autoencoder (VAE) branch to regularize the encoder. However, the implementation we use for this benchmark does not include the VAE branch.
\par SegResNet is a simple and robust architecture that works well with small sized datasets. However, since it is a CNN-based architecture,
it may have limitations in capturing long-range dependencies.
\\
\par We chose these four models for our benchmark because they all are commonly used when evaluating new architectures. Therefore, 
it is important to evaluate their performance independently to provide a fair comparison for future studies that use these models as baselines.

\subsection{Evaluation Metrics}
For evaluating the segmentation performance of the models, we calculated the Dice Similarity Coefficient (DSC) \cite{0000-11} for each organ. Then, We computed the mean DSC across all organs
to obtain an overall performance metric for each model. The DSC is defined as:
\begin{equation}
	DSC = \frac{2 |A \cap B|}{|A| + |B|}
\end{equation}
where A is the set of voxels in the predicted segmentation and B is the set of voxels in the ground truth segmentation. The DSC ranges from 0 to 1, with 1 indicating perfect overlap
between the predicted and ground truth segmentations.

\section{Results}
\subsection{Implementation Details}
All models except UNETR++ as well as the preprocessing, training, testing and validation steps were implemented using the MONAI \cite{0000-27} and PyTorch \cite{0000-28} Framework.
\par For UNETR we used a feature size of 16, the dimension of the hidden layer was 798, the dimension of the feedforward layer was 3072 and we used 12 
attention heads.
\par For SwinUNETR we used a feature size for 48, respectively the hidden layer dimensions were 48, 96, 192, 384 for the four stages of the Swin Transformer encoder and
we used 3, 6, 12, 24 attention heads for the four stages.
\par For SegResNet we used a initial filter size of 16 and the probability for dropout was 0.2. The number of downsample blocks in the four encoder layers were 1, 2, 2, 4 
and the number of upsample blocks in the three decoder layers were 1, 1, 1. 
The SegResNet implementation we use for this benchmark does not include the VAE branch.
\par For UNETR++ we used the official implementation provided by the authors \cite{0000-06} via this GitHub repository: \url{https://github.com/Amshaker/unetr_plus_plus}. 
We used a feature size of 16 as well as 4 attention heads and 3 EPA blocks on each of the four stages. The number of channels on each stage were
32, 64, 128 and 256.
\par We figured out that the hyperparameters as described above worked best for our dataset. A visualization of the architectures as they were used 
can be found in the appendix.
\par All models were trained on the same hardware, which consisted of a single NVIDIA A100 GPU with 40GB of VRAM. 
\subsection{Results and Discussion}
Table \ref{Tab-1} presents the results of the quantitative evaluation of the four models on the RATIC test set.
The table shows the mean DSC for each organ as well as the average DSC across all organs.
\par The CNN-based model SegResNet
outperforms the three transformer-based models UNETR, SwinUNETR and UNETR++ in terms of average DSC. It also achieves the highest DSC
for each individual organ. Among the transformer-based models, UNETR++ performs best. It demonstrates competitive performance, especially
for the left and right kidney as well as the liver, while slightly underperforming on segmenting the spleen and bowel. 
While UNETR shows a slightly worse but still strong performance, SwinUNETR performs significantly worse than the other models.
SwinUNETR is not just severely underperforming on segmenting the left kidney but also shows a significant drop in performance for all organs except 
the liver. This leads to a significantly lower average DSC for SwinUNETR compared to the others.
\par These findings indicate that Transformer-based models like UNETR, SwinUNETR and UNETER++ that previously claimed to outperform CNN-based models
like U-Net or SegResNet \cite{0000-01, 0000-04,0000-06} on segmentation tasks, do not necessarily achieve better performance than a well-optimized CNN-based model like 
SegResNet when trained on a medium-sized dataset of 144 scans. We assume that the reason for this is that CNNs possess a strong inductive bias 
toward local spatial context \cite{0000-16}. This makes them inherently more data-efficient than transformer-based models. 
\par These results support the findings of the study by Isensee et al. \cite{0000-23} which reports that when CNN-based architectures are properly configured, 
they often perform as well as, or better than, newer Transformer-based models. However, when comparing the performance of UNETR++ 
which was not part of the study by Isensee et al. and SegResNet, we see that the performance gap is minor with only 1.1\%. Since
the success of a model is often influenced by the specific dataset, its preprocessing as well as the hyperparameters used during training,
this margin is too small to draw a general conclusion about the superiority of one architecture over the other. Therefore we assume that a well-chosen and optimized transformer-based architecture can at least achieve similar performance as a CNN-based architecture.
\par
\begin{figure}[h]
	\centering
	\includegraphics[width=0.8\textwidth]{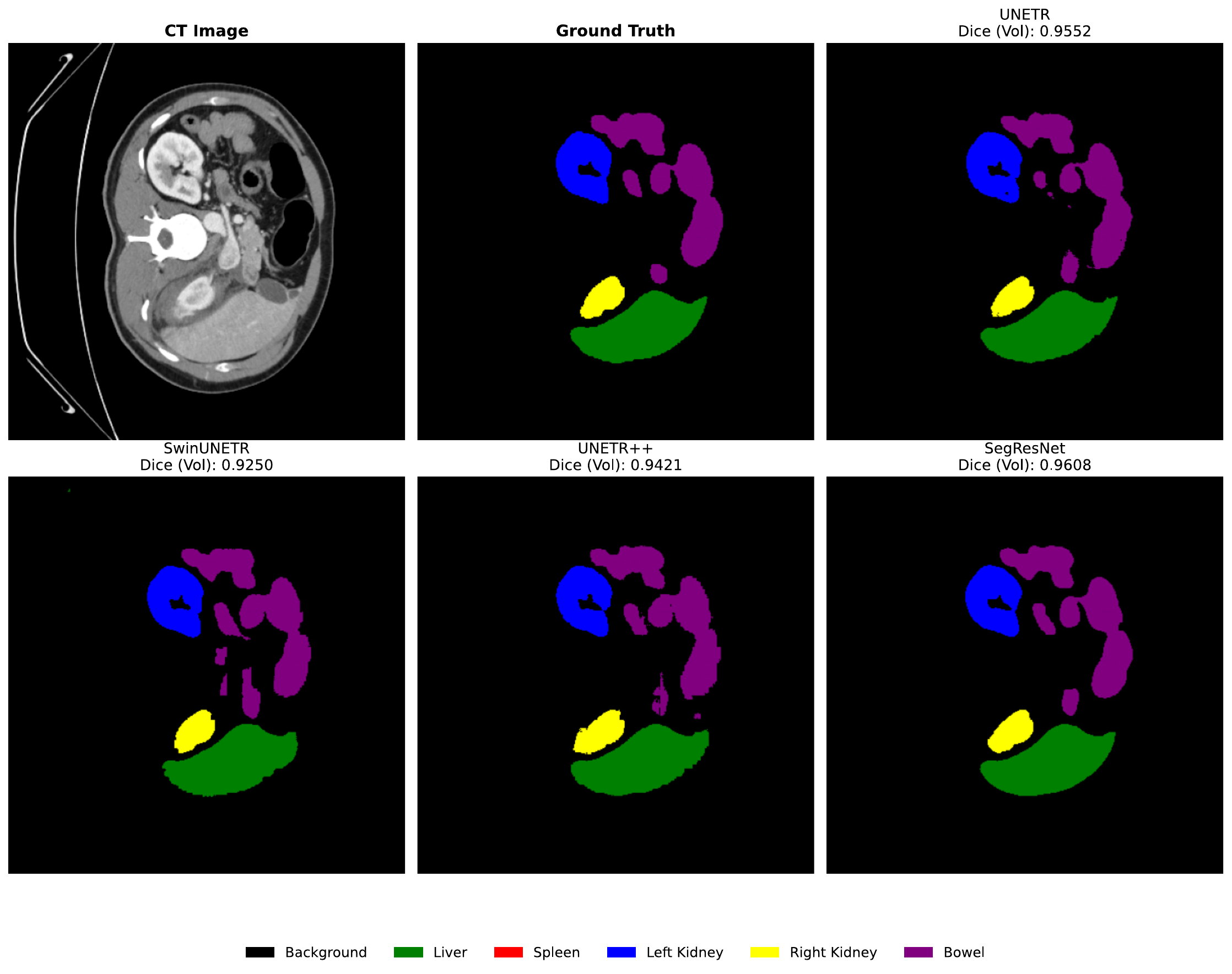}
	\caption{Qualitative comparison of the performance of UNETR, SwinUNETR, UNETR++ and SegResNet on a test scan from the RATIC dataset. 
	The figure shows the predicted segmentation masks for each model next to the original CT scan. 
	The ground truth segmentation is shown for comparison.}
	\label{Fig-2}
\end{figure}

 With the exception of SwinUNETR's outlier regarding the left kidney, table \ref{Tab-1} also shows that both CNN-based and 
transformer-based models achieve better performance on organs like liver or kidneys compared to the performance on spleen, bowel.
Therefore transformer-based as well as CNN-based models seem to have similar difficulties in learning the structures of non-uniform, more complex 
and disjointed organs like the bowel, while they perform better on more uniform and compact organs like the liver or kidneys. This can also be 
observed in Figure \ref{Fig-2}, which shows a qualitative comparison of the segmentation results of the four models on a test scan from the 
RATIC dataset.

\begin{table}[H]
	\centering
	\begin{tabular}{
		>{\raggedright\arraybackslash}m{2.5cm}*{6}{>{\centering\arraybackslash}m{1.4cm}}}
		\toprule
		          & Spleen         & Right Kidney   & Left Kidney    & Liver          & Bowel          & Average        \\
		\midrule
		UNETR     & 0.863          & 0.936          & 0.949          & 0.958          & 0.886          & 0.918          \\
		SwinUNETR & 0.811          & 0.876          & 0.698          & 0.933          & 0.826          & 0.829          \\
		UNETR++   & 0.907          & 0.945          & 0.959          & 0.963          & 0.902          & 0.934          \\
		\midrule
		SegResNet & \textbf{0.919} & \textbf{0.954} & \textbf{0.960} & \textbf{0.970} & \textbf{0.923} & \textbf{0.945} \\

		\bottomrule
	\end{tabular}
	\caption{Quantitative comparisons of segmentation performance on RATIC \cite{0000-15} test set.
		The table shows the mean Dice Similarity Coefficient (DSC) for Spleen, Right Kidney, Left Kidney and Liver, Bowel and the average
		DSC across all 5 organs in the RATIC dataset. The top section containing the transformer-based models and the bottom section
		containing the CNN-based model.}
	\label{Tab-1}
\end{table}

Table \ref{Tab-2} shows the number of training iterations needed for each model to achieve the results. The number
of iterations is the total number of batches processed over the entire training period. The batch size was 6 for all models.
\par The table shows that there is no significant difference between the CNN-based model and most transformer-based models. However,
UNETR++ requires only about half the number of iterations compared to the other models. This is likely due to the use of 
"Efficient Paired Attention" (EPA) in UNETR++, which combines spatial and channel attention to capture both local and global features 
more effectively. This allows UNETR++ to converge faster during training, even though its final performance is slightly lower than SegResNet.

\begin{table}[H]
	\centering
	\begin{tabular}{
		>{\raggedright\arraybackslash}m{2.5cm}*{6}{>{\centering\arraybackslash}m{1.4cm}}}
		\toprule
		          & Iterations    \\
		\midrule
		UNETR     & ~10,000       \\
		SwinUNETR & ~10,000       \\
		UNETR++   & \textbf{~5,000} \\
		\midrule
		SegResNet & ~10,000       \\

		\bottomrule
	\end{tabular}
	\caption{
		The table shows the number of training iterations needed for each model to achieve the results. The number
		of iterations is the total number of batches processed over the entire training period. The batch size
		was 6 for all models.
	}
	\label{Tab-2}
\end{table}

\section{Conclusion}
Our benchmark demonstrates that the CNN-based model SegResNet outperforms the three transformer-based models UNETR, SwinUNETR and UNETR++ in
abdominal multi-organ segmentation on the RATIC dataset. Therefore we conclude that, well-optimized CNN architectures remain highly competitive.
\par Furthermore our results reveal that the performance claims for abdominal multi-organ segmentation of some transformer-based architectures 
like SwinUNETR do not hold up when evaluated under independent and 
identical conditions. Which supports questions raised by Isensee et al. \cite{0000-23} about how currently new architectures are evaluated.
\par However, our results also show that the performance gap between the best transformer-based model UNETR++ and the CNN-based model SegResNet
is minor. Especially when also considering the resources needed for training, the models UNETR++ has an advantage. Therefore we consider 
certain transformer-based models like UNETR++ to be competitive.

\printbibliography

@misc{0000-01,
  title         = {UNETR: Transformers for 3D Medical Image Segmentation},
  author        = {Ali Hatamizadeh and Yucheng Tang and Vishwesh Nath and Dong Yang and Andriy Myronenko and Bennett Landman and Holger Roth and Daguang Xu},
  year          = {2021},
  eprint        = {2103.10504},
  archiveprefix = {arXiv},
  primaryclass  = {eess.IV},
  url           = {https://arxiv.org/abs/2103.10504}
}

@article{0000-02,
  author     = {Alexey Dosovitskiy and
                Lucas Beyer and
                Alexander Kolesnikov and
                Dirk Weissenborn and
                Xiaohua Zhai and
                Thomas Unterthiner and
                Mostafa Dehghani and
                Matthias Minderer and
                Georg Heigold and
                Sylvain Gelly and
                Jakob Uszkoreit and
                Neil Houlsby},
  title      = {An Image is Worth 16x16 Words: Transformers for Image Recognition
                at Scale},
  journal    = {CoRR},
  volume     = {abs/2010.11929},
  year       = {2020},
  url        = {https://arxiv.org/abs/2010.11929},
  eprinttype = {arXiv},
  eprint     = {2010.11929},
  bibsource  = {dblp computer science bibliography, https://dblp.org}
}

@article{0000-03,
  author     = {Olaf Ronneberger and
                Philipp Fischer and
                Thomas Brox},
  title      = {U-Net: Convolutional Networks for Biomedical Image Segmentation},
  journal    = {CoRR},
  volume     = {abs/1505.04597},
  year       = {2015},
  url        = {http://arxiv.org/abs/1505.04597},
  eprinttype = {arXiv},
  eprint     = {1505.04597},
  bibsource  = {dblp computer science bibliography, https://dblp.org}
}

@misc{0000-04,
  title         = {Swin UNETR: Swin Transformers for Semantic Segmentation of Brain Tumors in MRI Images},
  author        = {Ali Hatamizadeh and Vishwesh Nath and Yucheng Tang and Dong Yang and Holger Roth and Daguang Xu},
  year          = {2022},
  eprint        = {2201.01266},
  archiveprefix = {arXiv},
  primaryclass  = {eess.IV},
  url           = {https://arxiv.org/abs/2201.01266}
}

@misc{0000-05,
  title         = {Swin Transformer: Hierarchical Vision Transformer using Shifted Windows},
  author        = {Ze Liu and Yutong Lin and Yue Cao and Han Hu and Yixuan Wei and Zheng Zhang and Stephen Lin and Baining Guo},
  year          = {2021},
  eprint        = {2103.14030},
  archiveprefix = {arXiv},
  primaryclass  = {cs.CV},
  url           = {https://arxiv.org/abs/2103.14030}
}

@misc{0000-06,
  title         = {UNETR++: Delving into Efficient and Accurate 3D Medical Image Segmentation},
  author        = {Abdelrahman Shaker and Muhammad Maaz and Hanoona Rasheed and Salman Khan and Ming-Hsuan Yang and Fahad Shahbaz Khan},
  year          = {2024},
  eprint        = {2212.04497},
  archiveprefix = {arXiv},
  primaryclass  = {cs.CV},
  url           = {https://arxiv.org/abs/2212.04497}
}

@misc{0000-07,
  title         = {3D MRI brain tumor segmentation using autoencoder regularization},
  author        = {Andriy Myronenko},
  year          = {2018},
  eprint        = {1810.11654},
  archiveprefix = {arXiv},
  primaryclass  = {cs.CV},
  url           = {https://arxiv.org/abs/1810.11654}
}

@article{0000-08,
  author  = {Rudie, Jeffrey D. and Lin, Hui-Ming and Ball, Robyn L. and Jalal, Sabeena and Prevedello, Luciano M. and Nicolaou, Savvas and Marinelli, Brett S. and Flanders, Adam E. and Magudia, Kirti and Shih, George and Davis, Melissa A. and Mongan, John and Chang, Peter D. and Berger, Ferco H. and Hermans, Sebastiaan and Law, Meng and Richards, Tyler and Grunz, Jan-Peter and Kunz, Andreas Steven and Mathur, Shobhit and Galea-Soler, Sandro and Chung, Andrew D. and Afat, Saif and Kuo, Chin-Chi and Aweidah, Layal},
  title   = {The RSNA Abdominal Traumatic Injury CT (RATIC) Dataset},
  journal = {Radiology: Artificial Intelligence},
  year    = {2024},
  volume  = {6},
  number  = {6},
  pages   = {e240101},
  doi     = {10.1148/ryai.240101}
}

@misc{0000-09,
  title         = {nnU-Net: Self-adapting Framework for U-Net-Based Medical Image Segmentation},
  author        = {Fabian Isensee and Jens Petersen and Andre Klein and David Zimmerer and Paul F. Jaeger and Simon Kohl and Jakob Wasserthal and Gregor Koehler and Tobias Norajitra and Sebastian Wirkert and Klaus H. Maier-Hein},
  year          = {2018},
  eprint        = {1809.10486},
  archiveprefix = {arXiv},
  primaryclass  = {cs.CV},
  url           = {https://arxiv.org/abs/1809.10486}
}

@article{0000-10,
  title     = {TotalSegmentator: Robust Segmentation of 104 Anatomic Structures in CT Images},
  volume    = {5},
  issn      = {2638-6100},
  url       = {http://dx.doi.org/10.1148/ryai.230024},
  doi       = {10.1148/ryai.230024},
  number    = {5},
  journal   = {Radiology: Artificial Intelligence},
  publisher = {Radiological Society of North America (RSNA)},
  author    = {Wasserthal, Jakob and Breit, Hanns-Christian and Meyer, Manfred T. and Pradella, Maurice and Hinck, Daniel and Sauter, Alexander W. and Heye, Tobias and Boll, Daniel T. and Cyriac, Joshy and Yang, Shan and Bach, Michael and Segeroth, Martin},
  year      = {2023},
  month     = sep
}

@article{0000-11,
  author  = {Dice, Lee R.},
  title   = {Measures of the Amount of Ecologic Association Between Species},
  journal = {Ecology},
  year    = {1945},
  volume  = {26},
  number  = {3},
  pages   = {297--302},
  doi     = {10.2307/1932409}
}

@inproceedings{0000-15,
  title        = {Miccai multi-atlas labeling beyond the cranial vault--workshop and challenge},
  author       = {Landman, Bennett and Xu, Zhoubing and Igelsias, Juan and Styner, Martin and Langerak, Thomas and Klein, Arno},
  booktitle    = {Proc. MICCAI multi-atlas labeling beyond cranial vault—workshop challenge},
  volume       = {5},
  pages        = {12},
  year         = {2015},
  organization = {Munich, Germany}
}

@misc{0000-16,
  title         = {An Image is Worth 16x16 Words: Transformers for Image Recognition at Scale},
  author        = {Alexey Dosovitskiy and Lucas Beyer and Alexander Kolesnikov and Dirk Weissenborn and Xiaohua Zhai and Thomas Unterthiner and Mostafa Dehghani and Matthias Minderer and Georg Heigold and Sylvain Gelly and Jakob Uszkoreit and Neil Houlsby},
  year          = {2021},
  eprint        = {2010.11929},
  archiveprefix = {arXiv},
  primaryclass  = {cs.CV},
  url           = {https://arxiv.org/abs/2010.11929}
}

@inproceedings{0000-17,
  author    = {Li, QingFeng and Tong, Jigang and Yang, Sen and Du, Shengzhi},
  booktitle = {2024 IEEE International Conference on Mechatronics and Automation (ICMA)},
  title     = {FATUnetr:fully attention Transformer for 3D medical image segmentation},
  year      = {2024},
  volume    = {},
  number    = {},
  pages     = {1415-1419},
  doi       = {10.1109/ICMA61710.2024.10633071}
}

@misc{0000-18,
  title         = {SegResMamba: An Efficient Architecture for 3D Medical Image Segmentation},
  author        = {Badhan Kumar Das and Ajay Singh and Saahil Islam and Gengyan Zhao and Andreas Maier},
  year          = {2025},
  eprint        = {2503.07766},
  archiveprefix = {arXiv},
  primaryclass  = {cs.CV},
  url           = {https://arxiv.org/abs/2503.07766}
}

@article{0000-19,
  author   = {Xing, Zhaohu and Zhu, Lei and Yu, Lequan and Xing, Zhiheng and Wan, Liang},
  journal  = {IEEE Journal of Biomedical and Health Informatics},
  title    = {Hybrid Masked Image Modeling for 3D Medical Image Segmentation},
  year     = {2024},
  volume   = {28},
  number   = {4},
  pages    = {2115-2125},
  doi      = {10.1109/JBHI.2024.3360239}
}

@inproceedings{0000-20,
  author    = {Tang, Yucheng and Yang, Dong and Li, Wenqi and Roth, Holger R. and Landman, Bennett and Xu, Daguang and Nath, Vishwesh and Hatamizadeh, Ali},
  title     = {Self-Supervised Pre-Training of Swin Transformers for 3D Medical Image Analysis},
  booktitle = {Proceedings of the IEEE/CVF Conference on Computer Vision and Pattern Recognition (CVPR)},
  month     = {7},
  year      = {2022},
  pages     = {20730-20740}
}

@misc{0000-21,
  title     = {Segmentation Outside the Cranial Vault Challenge},
  url       = {https://repo-prod.prod.sagebase.org/repo/v1/doi/locate?id=syn3193805&type=ENTITY},
  doi       = {10.7303/SYN3193805},
  publisher = {Synapse},
  author    = {harrigr},
  year      = {2015}
}

@InProceedings{0000-22,
author="Chang, Ao
and Zeng, Jiajun
and Huang, Ruobing
and Ni, Dong",
editor="Linguraru, Marius George
and Dou, Qi
and Feragen, Aasa
and Giannarou, Stamatia
and Glocker, Ben
and Lekadir, Karim
and Schnabel, Julia A.",
title="EM-Net: Efficient Channel and Frequency Learning with Mamba for 3D Medical Image Segmentation",
booktitle="Medical Image Computing and Computer Assisted Intervention -- MICCAI 2024",
year="2024",
publisher="Springer Nature Switzerland",
address="Cham",
pages="266--275",
isbn="978-3-031-72114-4"
}

@InProceedings{0000-23,
author="Isensee, Fabian
and Wald, Tassilo
and Ulrich, Constantin
and Baumgartner, Michael
and Roy, Saikat
and Maier-Hein, Klaus
and J{\"a}ger, Paul F.",
editor="Linguraru, Marius George
and Dou, Qi
and Feragen, Aasa
and Giannarou, Stamatia
and Glocker, Ben
and Lekadir, Karim
and Schnabel, Julia A.",
title="nnU-Net Revisited: A Call for Rigorous Validation in 3D Medical Image Segmentation",
booktitle="Medical Image Computing and Computer Assisted Intervention -- MICCAI 2024",
year="2024",
publisher="Springer Nature Switzerland",
address="Cham",
pages="488--498",
}

@article{0000-25,
  author = {Takahashi, Satoshi and Sakaguchi, Yusuke and Kouno, Nobuji and Takasawa, Ken and Ishizu, Kenichi and Akagi, Yu and Aoyama, Rina and Teraya, Naoki and Bolatkan, Amina and Shinkai, Norio and Machino, Hidenori and Kobayashi, Kazuma und Asada, Ken and Komatsu, Masaaki and Kaneko, Syuzo and Sugiyama, Masashi and Hamamoto, Ryuji},
  title = {Comparison of Vision Transformers and Convolutional Neural Networks in Medical Image Analysis: A Systematic Review},
  journal = {Journal of Medical Systems},
  year = {2024},
  volume = {48},
  number = {1},
  pages = {84},
  month = {Sep},
  doi = {10.1007/s10916-024-02105-8},
  publisher = {Springer Nature},
  url = {https://doi.org/10.1007/s10916-024-02105-8}
}

@article{0000-26,
  author = {Pu, Qian and Xi, Zhenya and Yin, Shuzhen and Zhao, Zhanjun and Zhao, Li},
  title = {Advantages of transformer and its application for medical image segmentation: a survey},
  journal = {BioMedical Engineering OnLine},
  year = {2024},
  volume = {23},
  number = {1},
  pages = {14},
  doi = {10.1186/s12938-024-01212-4},
  url = {https://doi.org/10.1186/s12938-024-01212-4}
}

@misc{0000-27,
      title={MONAI: An open-source framework for deep learning in healthcare}, 
      author={M. Jorge Cardoso and Wenqi Li and Richard Brown and Nic Ma and Eric Kerfoot and Yiheng Wang and Benjamin Murrey and Andriy Myronenko and Can Zhao and Dong Yang and Vishwesh Nath and Yufan He and Ziyue Xu and Ali Hatamizadeh and Andriy Myronenko and Wentao Zhu and Yun Liu and Mingxin Zheng and Yucheng Tang and Isaac Yang and Michael Zephyr and Behrooz Hashemian and Sachidanand Alle and Mohammad Zalbagi Darestani and Charlie Budd and Marc Modat and Tom Vercauteren and Guotai Wang and Yiwen Li and Yipeng Hu and Yunguan Fu and Benjamin Gorman and Hans Johnson and Brad Genereaux and Barbaros S. Erdal and Vikash Gupta and Andres Diaz-Pinto and Andre Dourson and Lena Maier-Hein and Paul F. Jaeger and Michael Baumgartner and Jayashree Kalpathy-Cramer and Mona Flores and Justin Kirby and Lee A. D. Cooper and Holger R. Roth and Daguang Xu and David Bericat and Ralf Floca and S. Kevin Zhou and Haris Shuaib and Keyvan Farahani and Klaus H. Maier-Hein and Stephen Aylward and Prerna Dogra and Sebastien Ourselin and Andrew Feng},
      year={2022},
      eprint={2211.02701},
      archivePrefix={arXiv},
      primaryClass={cs.LG},
      url={https://arxiv.org/abs/2211.02701}, 
}

@incollection{0000-28,
title = {PyTorch: An Imperative Style, High-Performance Deep Learning Library},
author = {Paszke, Adam and Gross, Sam and Massa, Francisco and Lerer, Adam and Bradbury, James and Chanan, Gregory and Killeen, Trevor and Lin, Zeming and Gimelshein, Natalia and Antiga, Luca and Desmaison, Alban and Kopf, Andreas and Yang, Edward and DeVito, Zachary and Raison, Martin and Tejani, Alykhan and Chilamkurthy, Sasank and Steiner, Benoit and Fang, Lu and Bai, Junjie and Chintala, Soumith},
booktitle = {Advances in Neural Information Processing Systems 32},
pages = {8024--8035},
year = {2019},
publisher = {Curran Associates, Inc.},
url = {http://papers.neurips.cc/paper/9015-pytorch-an-imperative-style-high-performance-deep-learning-library.pdf}
}

\newpage
\section{Appendix}
\subsection{Architectures}
Figure \ref{Fig-3}, \ref{Fig-4}, \ref{Fig-5} and \ref{Fig-6} show schematic visualizations of the UNETR, 
SwinUNETR, UNETR++ and SegResNet architectures as used in this benchmark.
\par For UNETR, SwinUNETR and SegResNet we used the MONAI \cite{0000-27} implementations, which are accessible via its GitHub 
repository: \url{https://github.com/Project-MONAI/MONAI}. For UNETR++ we used the official implementation
provided by the authors \cite{0000-06} via this github repository: \url{https://github.com/Amshaker/unetr_plus_plus}.
The complete code for data preprocessing, training, testing and validation as well as the implementation of the models
for this paper is available on github: \url{https://github.com/lukas-FAU/medical_image_segmentation_benchmarking_paper}.

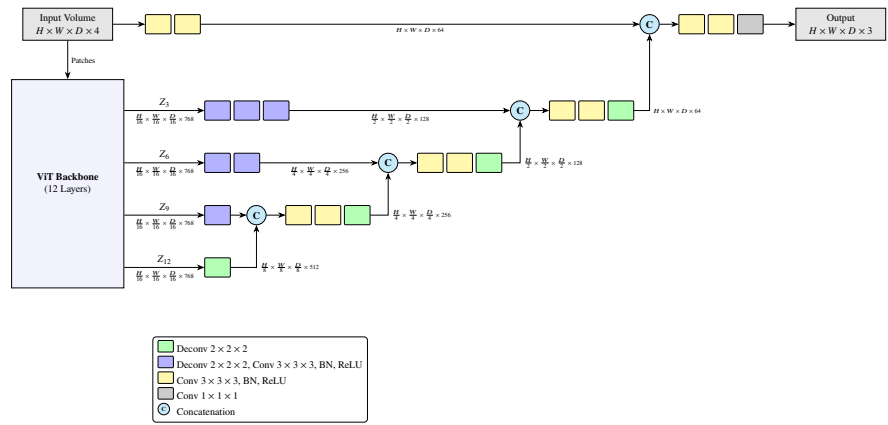
\begin{figure}[ht]
	\centering
	\resizebox{\textwidth}{!}{
    \begin{tikzpicture}[
            node distance=0.4cm,
            volume/.style={draw, fill=gray!20, minimum width=2.8cm, minimum height=1cm, align=center, thick},
            vit/.style={draw, fill=blue!5, minimum width=3.5cm, minimum height=6.5cm, align=center, thick},
            bluebox/.style={draw, fill=blue!30, minimum width=0.8cm, minimum height=0.6cm, rounded corners=1pt},
            greenbox/.style={draw, fill=green!30, minimum width=0.8cm, minimum height=0.6cm, rounded corners=1pt},
            yellowbox/.style={draw, fill=yellow!40, minimum width=0.8cm, minimum height=0.6cm, rounded corners=1pt},
            greybox/.style={draw, fill=gray!40, minimum width=0.8cm, minimum height=0.6cm, rounded corners=1pt},
            circle_c/.style={draw, fill=cyan!20, circle, minimum size=0.6cm, inner sep=0pt, font=\small\bfseries},
            arrow/.style={-Stealth, thick},
            dimlabel/.style={font=\tiny, align=center}
        ]

        \node[volume] (input) {Input Volume \\ $H \times W \times D \times 4$};

        \node[yellowbox, right=1cm of input] (iy1) {};
        \node[yellowbox, right=0.1cm of iy1] (iy2) {};
        \draw[arrow] (input) -- (iy1);

        \node[vit, below=1.2cm of input] (vit) {\textbf{ViT Backbone} \\ (12 Layers)};
        \draw[arrow] (input) -- node[right, font=\scriptsize] {Patches} (vit);

        \node[circle_c] (concat_final) at ($(iy2.east) + (14, 0)$) {C};
        \draw[arrow] (iy2.east) -- node[below, midway, dimlabel] {$H \times W \times D \times 64$}(concat_final.west);


        \coordinate (z3_start) at ($(vit.north east)!0.15!(vit.south east)$);
        \node[bluebox, right=2.5cm of z3_start] (z3_b1) {};
        \node[bluebox, right=0.1cm of z3_b1] (z3_b2) {};
        \node[bluebox, right=0.1cm of z3_b2] (z3_b3) {};
        \draw[arrow] (z3_start) -- node[above, midway, font=\small] {$Z_3$}
        node[below, midway, dimlabel] {$\frac{H}{16} \times \frac{W}{16} \times \frac{D}{16} \times 768$} (z3_b1);
        \node[circle_c] (concat3) at ($(z3_b3.east) + (7.2, 0)$) {C};
        \draw[arrow] (z3_b3.east) -- node[below, midway, dimlabel] {$\frac{H}{2} \times \frac{W}{2} \times \frac{D}{2} \times 128$} (concat3.west);
        \node[yellowbox, right=0.6cm of concat3] (y3_1) {};
        \node[yellowbox, right=0.1cm of y3_1] (y3_2) {};
        \node[greenbox, right=0.1cm of y3_2] (g3_final) {};
        \draw[arrow] (concat3.east) -- (y3_1);

        \coordinate (z6_start) at ($(vit.north east)!0.4!(vit.south east)$);
        \node[bluebox, right=2.5cm of z6_start] (z6_b1) {};
        \node[bluebox, right=0.1cm of z6_b1] (z6_b2) {};
        \draw[arrow] (z6_start) -- node[above, midway, font=\small] {$Z_6$}
        node[below, midway, dimlabel] {$\frac{H}{16} \times \frac{W}{16} \times \frac{D}{16} \times 768$} (z6_b1);
        \node[circle_c] (concat6) at ($(z6_b2.east) + (4.0, 0)$) {C};
        \draw[arrow] (z6_b2.east) -- node[below, midway, dimlabel] {$\frac{H}{4} \times \frac{W}{4} \times \frac{D}{4} \times 256$} (concat6.west);
        \node[yellowbox, right=0.6cm of concat6] (y6_1) {};
        \node[yellowbox, right=0.1cm of y6_1] (y6_2) {};
        \node[greenbox, right=0.1cm of y6_2] (g6_final) {};
        \draw[arrow] (concat6.east) -- (y6_1);

        \coordinate (z9_start) at ($(vit.north east)!0.65!(vit.south east)$);
        \node[bluebox, right=2.5cm of z9_start] (z9_b1) {};
        \node[circle_c, right=0.5cm of z9_b1] (concat9) {C};
        \draw[arrow] (z9_start) -- node[above, midway, font=\small] {$Z_9$}
        node[below, midway, dimlabel] {$\frac{H}{16} \times \frac{W}{16} \times \frac{D}{16} \times 768$} (z9_b1);
        \draw[arrow] (z9_b1.east) -- (concat9.west);

        \coordinate (z12_start) at ($(vit.north east)!0.9!(vit.south east)$);
        \node[greenbox, right=2.5cm of z12_start] (z12_g1) {};
        \draw[arrow] (z12_start) -- node[above, midway, font=\small] {$Z_{12}$}
        node[below, midway, dimlabel] {$\frac{H}{16} \times \frac{W}{16} \times \frac{D}{16} \times 768$} (z12_g1);
        \draw[arrow] (z12_g1.east) -- ++(0.5,0) -| node[right, midway, dimlabel] {$\frac{H}{8} \times \frac{W}{8} \times \frac{D}{8} \times 512$}  (concat9.south);

        \node[yellowbox, right=0.6cm of concat9] (y9_1) {};
        \node[yellowbox, right=0.1cm of y9_1] (y9_2) {};
        \node[greenbox, right=0.1cm of y9_2] (g9_final) {};
        \draw[arrow] (concat9.east) -- (y9_1);

        \draw[arrow] (g9_final.east) -- ++(0.5,0) -| node[right, midway, dimlabel] {$\frac{H}{4} \times \frac{W}{4} \times \frac{D}{4} \times 256$} (concat6.south);
        \draw[arrow] (g6_final.east) -- ++(0.5,0) -| node[right, midway, dimlabel] {$\frac{H}{2} \times \frac{W}{2} \times \frac{D}{2} \times 128$} (concat3.south);
        \draw[arrow] (g3_final.east) -- ++(0.5,0) -| node[right, midway, dimlabel] {$ H \times W \times D \times 64$} (concat_final.south);

        \node[yellowbox, right=0.6cm of concat_final] (yf1) {};
        \node[yellowbox, right=0.1cm of yf1] (yf2) {};
        \node[greybox, right=0.1cm of yf2] (gf) {};
        \node[volume, right=1cm of gf] (output) {Output \\ $H \times W \times D \times 3$};
        \draw[arrow] (concat_final.east) -- (yf1);
        \draw[arrow] (gf.east) -- (output);

        \node[draw, thick, fill=white, below=1.5cm of vit.south, xshift=6cm, rounded corners=3pt] (legend) {
            \begin{tabular}{ll}
                \tikz\node[greenbox, scale=0.6]{};  & Deconv $2 \times 2 \times 2$                                       \\
                \tikz\node[bluebox, scale=0.6]{};   & Deconv $2 \times 2 \times 2$, Conv $3 \times 3 \times 3$, BN, ReLU \\
                \tikz\node[yellowbox, scale=0.6]{}; & Conv $3 \times 3 \times 3$, BN, ReLU                               \\
                \tikz\node[greybox, scale=0.6]{};   & Conv $1 \times 1 \times 1$                                         \\
                \tikz\node[circle_c, scale=0.6]{C}; & Concatenation
            \end{tabular}
        };

    \end{tikzpicture}
}
	\caption{Schematic visualization of the UNETR architecture as used in this benchmark. \cite{0000-01}}
	\label{Fig-3}
\end{figure}
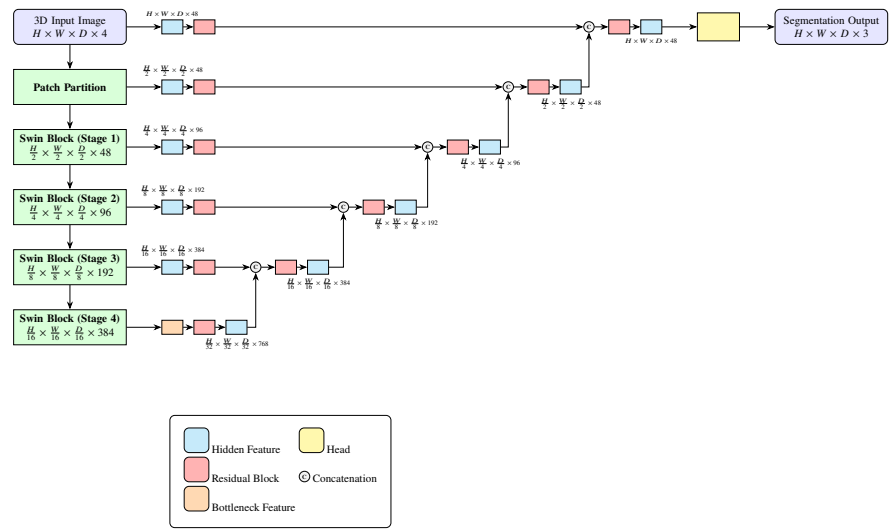
\begin{figure}[ht]
	\centering
	\resizebox{\textwidth}{!}{

    \begin{tikzpicture}[
            node distance=0.7cm and 1.2cm,
            auto,
            block/.style={rectangle, draw, fill=blue!10, text width=3cm, align=center, minimum height=1cm, rounded corners},
            swin/.style={rectangle, draw, fill=green!15, text width=3cm, align=center, minimum height=1cm, font=\small\bfseries},
            feat_orange/.style={rectangle, draw, fill=orange!30, minimum width=0.6cm, minimum height=0.4cm},
            feat_purple/.style={rectangle, draw, fill=purple!20, minimum width=0.6cm, minimum height=0.4cm},
            feat_cyan/.style={rectangle, draw, fill=cyan!20,   minimum width=0.6cm, minimum height=0.4cm},
            feat_red/.style={rectangle, draw, fill=red!30,    minimum width=0.6cm, minimum height=0.4cm},
            head_block/.style={rectangle, draw, fill=yellow!40, minimum width=1.2cm, minimum height=0.8cm},
            concat/.style={circle, draw, inner sep=1.2pt, font=\tiny\bfseries, fill=gray!10},
            arrow/.style={-Stealth, thick}
        ]

        \node (input)  [block] {3D Input Image \\$H\times W\times D\times 4$};
        \node (patch)  [swin,  below=of input] {Patch Partition};
        \node (stage1) [swin,  below=of patch] {Swin Block (Stage 1) \\ $\frac{H}{2}\times\frac{W}{2}\times\frac{D}{2}\times48$};
        \node (stage2) [swin,  below=of stage1] {Swin Block (Stage 2) \\ $\frac{H}{4}\times\frac{W}{4}\times\frac{D}{4}\times96$};
        \node (stage3) [swin,  below=of stage2] {Swin Block (Stage 3) \\ $\frac{H}{8}\times\frac{W}{8}\times\frac{D}{8}\times192$};
        \node (stage4) [swin,  below=of stage3] {Swin Block (Stage 4) \\ $\frac{H}{16}\times\frac{W}{16}\times\frac{D}{16}\times384$};


        \node (s4_bneck) [feat_orange, right=1cm of stage4] {};
        \node (s4_res)   [feat_red, right=0.3cm of s4_bneck] {};
        \node (s4_hid)   [feat_cyan,   right=0.3cm of s4_res, label={[font=\tiny]below:$\frac{H}{32}\times\frac{W}{32}\times\frac{D}{32}\times 768$}] {};

        \node (s3_hid)   [feat_cyan,   right=1cm of stage3, label={[font=\tiny]above:$\frac{H}{16}\times\frac{W}{16}\times\frac{D}{16}\times 384$}] {};
        \node (s3_res)   [feat_red,    right=0.3cm of s3_hid] {};
        \node (c_s3)     [concat,      right=1.0cm of s3_res] {C};
        \node (p3_res)   [feat_red,    right=0.4cm of c_s3] {};
        \node (p3_hid)   [feat_cyan,   right=0.3cm of p3_res, label={[font=\tiny]below:$\frac{H}{16}\times\frac{W}{16}\times\frac{D}{16}\times 384$}] {};

        \node (s2_hid)   [feat_cyan,   right=1cm of stage2, label={[font=\tiny]above:$\frac{H}{8}\times\frac{W}{8}\times\frac{D}{8}\times 192$}] {};
        \node (s2_res)   [feat_red,    right=0.3cm of s2_hid] {};
        \node (c_s2)     [concat,      right=3.5cm of s2_res] {C};
        \node (p2_res)   [feat_red,    right=0.4cm of c_s2] {};
        \node (p2_hid)   [feat_cyan,   right=0.3cm of p2_res, label={[font=\tiny]below:$\frac{H}{8}\times\frac{W}{8}\times\frac{D}{8}\times 192$}] {};

        \node (s1_hid)   [feat_cyan,   right=1cm of stage1, label={[font=\tiny]above:$\frac{H}{4}\times\frac{W}{4}\times\frac{D}{4}\times 96$}] {};
        \node (s1_res)   [feat_red,    right=0.3cm of s1_hid] {};
        \node (c_s1)     [concat,      right=5.9cm of s1_res] {C};
        \node (p1_res)   [feat_red,    right=0.4cm of c_s1] {};
        \node (p1_hid)   [feat_cyan,   right=0.3cm of p1_res, label={[font=\tiny]below:$\frac{H}{4}\times\frac{W}{4}\times\frac{D}{4}\times 96$}] {};

        \node (sp_hid)   [feat_cyan,   right=1cm of patch, label={[font=\tiny]above:$\frac{H}{2}\times\frac{W}{2}\times\frac{D}{2}\times 48$}] {};
        \node (sp_res)   [feat_red,    right=0.3cm of sp_hid] {};
        \node (c_sp)     [concat,      right=8.2cm of sp_res] {C};
        \node (pp_res)   [feat_red,    right=0.4cm of c_sp] {};
        \node (pp_hid)   [feat_cyan,   right=0.3cm of pp_res, label={[font=\tiny]below:$\frac{H}{2}\times\frac{W}{2}\times\frac{D}{2}\times 48$}] {};

        \node (si_hid)   [feat_cyan,   right=1cm of input, label={[font=\tiny]above:$H\times W\times D\times 48$}] {};
        \node (si_res)   [feat_red,    right=0.3cm of si_hid] {};
        \node (c_si)     [concat,      right=10.5cm of si_res] {C};
        \node (pi_res)   [feat_red,    right=0.4cm of c_si] {};
        \node (pi_hid)   [feat_cyan,   right=0.3cm of pi_res, label={[font=\tiny]below:$H\times W\times D\times 48$}] {};

        \node (head)     [head_block, right=1cm of pi_hid] {};
        \node (output)   [block, right=1cm of head] {Segmentation Output \\ $H\times W\times D\times 3$};

        \draw [arrow] (input) -- (patch);
        \draw [arrow] (patch) -- (stage1);
        \draw [arrow] (stage1) -- (stage2);
        \draw [arrow] (stage2) -- (stage3);
        \draw [arrow] (stage3) -- (stage4);

        \draw [arrow] (stage4) -- (s4_bneck); \draw [arrow] (s4_bneck) -- (s4_res); \draw [arrow] (s4_res) -- (s4_hid);
        \draw [arrow] (s4_hid.east) -| (c_s3.south);

        \draw [arrow] (stage3) -- (s3_hid); \draw [arrow] (s3_hid) -- (s3_res); \draw [arrow] (s3_res) -- (c_s3);
        \draw [arrow] (c_s3) -- (p3_res); \draw [arrow] (p3_res) -- (p3_hid);
        \draw [arrow] (p3_hid.east) -| (c_s2.south);

        \draw [arrow] (stage2) -- (s2_hid); \draw [arrow] (s2_hid) -- (s2_res); \draw [arrow] (s2_res) -- (c_s2);
        \draw [arrow] (c_s2) -- (p2_res); \draw [arrow] (p2_res) -- (p2_hid);
        \draw [arrow] (p2_hid.east) -| (c_s1.south);

        \draw [arrow] (stage1) -- (s1_hid); \draw [arrow] (s1_hid) -- (s1_res); \draw [arrow] (s1_res) -- (c_s1);
        \draw [arrow] (c_s1) -- (p1_res); \draw [arrow] (p1_res) -- (p1_hid);
        \draw [arrow] (p1_hid.east) -| (c_sp.south);

        \draw [arrow] (patch) -- (sp_hid); \draw [arrow] (sp_hid) -- (sp_res); \draw [arrow] (sp_res) -- (c_sp);
        \draw [arrow] (c_sp) -- (pp_res); \draw [arrow] (pp_res) -- (pp_hid);
        \draw [arrow] (pp_hid.east) -| (c_si.south);

        \draw [arrow] (input) -- (si_hid); \draw [arrow] (si_hid) -- (si_res); \draw [arrow] (si_res) -- (c_si);
        \draw [arrow] (c_si) -- (pi_res); \draw [arrow] (pi_res) -- (pi_hid);

        \draw [arrow] (pi_hid) -- (head);
        \draw [arrow] (head) -- (output);

        \node [below=2cm of stage4, xshift=6cm, draw, inner sep=10pt, rounded corners, fill=white] (legend) {
            \begin{tabular}{ll}
                \tikz\node[feat_cyan, minimum width=0.4cm, minimum height=0.3cm]{}; Hidden Feature       & \tikz\node[head_block, minimum width=0.4cm, minimum height=0.3cm]{}; Head \\
                \tikz\node[feat_red, minimum width=0.4cm, minimum height=0.3cm]{}; Residual Block        & \tikz\node[concat, minimum width=0.3cm]{C}; Concatenation                 \\
                \tikz\node[feat_orange, minimum width=0.4cm, minimum height=0.3cm]{}; Bottleneck Feature &                                                                           \\
            \end{tabular}
        };

    \end{tikzpicture}
}
	\caption{Schematic visualization of the SwinUNETR architecture as used in this benchmark. \cite{0000-04}}
	\label{Fig-4}
\end{figure}
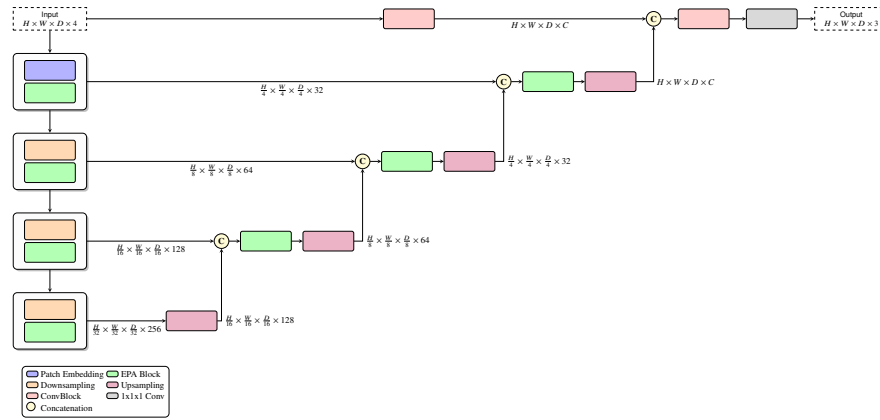
\begin{figure}[ht]
	\centering
	\resizebox{\textwidth}{!}{

    \begin{tikzpicture}[
            node distance=0.8cm,
            base/.style={draw, thick, rounded corners, minimum width=2.6cm, minimum height=2.0cm, align=center, fill=gray!2, drop shadow},
            subbox/.style={draw, minimum width=1.8cm, minimum height=0.7cm, rounded corners=2pt},
            io_style/.style={draw, dashed, thick, minimum width=2.6cm, minimum height=0.8cm, align=center, font=\footnotesize\sffamily},
            concat/.style={draw, circle, inner sep=2pt, font=\small\bfseries, fill=yellow!20},
            patch_color/.style={fill=blue!30},
            epa_color/.style={fill=green!30},
            down_color/.style={fill=orange!30},
            up_color/.style={fill=purple!30},
            conv_color/.style={fill=red!20},
            final_conv_color/.style={fill=gray!30},
            arrow/.style={-stealth, thick},
            dimlabel/.style={font=\small\sffamily}
        ]

        \node (input) [io_style] {Input \\ $H \times W \times D \times 4$};

        \node (block1) [base, below=of input] {};
        \node at ($(block1.center) + (0,0.4)$) [subbox, patch_color] {};
        \node at ($(block1.center) + (0,-0.4)$) [subbox, epa_color] {};

        \node (block2) [base, below=of block1] {};
        \node at ($(block2.center) + (0,0.4)$) [subbox, down_color] {};
        \node at ($(block2.center) + (0,-0.4)$) [subbox, epa_color] {};

        \node (block3) [base, below=of block2] {};
        \node at ($(block3.center) + (0,0.4)$) [subbox, down_color] {};
        \node at ($(block3.center) + (0,-0.4)$) [subbox, epa_color] {};

        \node (block4) [base, below=of block3] {};
        \node at ($(block4.center) + (0,0.4)$) [subbox, down_color] {};
        \node at ($(block4.center) + (0,-0.4)$) [subbox, epa_color] {};

        \node (input_conv) [subbox, conv_color, right=10.5cm of input] {};
        \node (c0) [concat, right=7.5cm of input_conv] {C};
        \node (post_conv) [subbox, conv_color, right=0.6cm of c0] {};
        \node (final_conv) [subbox, final_conv_color, right=0.6cm of post_conv] {};
        \node (output_final) [io_style, right=0.6cm of final_conv] {Output \\ $H \times W \times D \times 3$};


        \node (up_bot) [subbox, up_color, right=2.8cm of block4] {};

        \node (c3) [concat, right=4.5cm of block3] {C};
        \node (e3) [subbox, epa_color, right=0.4cm of c3] {};
        \node (up3) [subbox, up_color, right=0.4cm of e3] {};

        \node (c2) [concat, right=9.5cm of block2] {C};
        \node (e2) [subbox, epa_color, right=0.4cm of c2] {};
        \node (up2) [subbox, up_color, right=0.4cm of e2] {};

        \node (c1) [concat, right=14.5cm of block1] {C};
        \node (e1) [subbox, epa_color, right=0.4cm of c1] {};
        \node (up_final) [subbox, up_color, right=0.4cm of e1] {};

        \draw [arrow] (input) -- (block1);
        \draw [arrow] (block1) -- (block2);
        \draw [arrow] (block2) -- (block3);
        \draw [arrow] (block3) -- (block4);

        \draw [arrow] (input.east) -- (input_conv.west);
        \draw [arrow] (block4.east) --node[below, midway, dimlabel] {$\frac{H}{32} \times \frac{W}{32} \times \frac{D}{32} \times 256$} (up_bot.west);

        \draw [arrow] (up_bot.east) -| node[right, midway, dimlabel] {$\frac{H}{16} \times \frac{W}{16} \times \frac{D}{16} \times 128$} (c3.south);
        \draw [arrow] (up3.east) -| node[right, midway, dimlabel] {$\frac{H}{8} \times \frac{W}{8} \times \frac{D}{8} \times 64$} (c2.south);
        \draw [arrow] (up2.east) -| node[right, midway, dimlabel] {$\frac{H}{4} \times \frac{W}{4} \times \frac{D}{4} \times 32$} (c1.south);

        \draw [arrow] (up_final.east) -| node[right, midway, dimlabel] {$H \times W \times D \times C$} (c0.south);

        \draw [arrow] (input_conv) -- node[below, midway, dimlabel] {$H \times W \times D \times C$}  (c0);
        \draw [arrow] (c0) -- (post_conv);
        \draw [arrow] (post_conv) -- (final_conv);
        \draw [arrow] (final_conv) -- (output_final);

        \draw [arrow] (c3) -- (e3); \draw [arrow] (e3) -- (up3);
        \draw [arrow] (c2) -- (e2); \draw [arrow] (e2) -- (up2);
        \draw [arrow] (c1) -- (e1); \draw [arrow] (e1) -- (up_final);

        \draw [arrow] (block3.east) -- node[below, midway, dimlabel] {$\frac{H}{16} \times \frac{W}{16} \times \frac{D}{16} \times 128$} (c3.west);
        \draw [arrow] (block2.east) -- node[below, midway, dimlabel] {$\frac{H}{8} \times \frac{W}{8} \times \frac{D}{8} \times 64$} (c2.west);
        \draw [arrow] (block1.east) -- node[below, midway, dimlabel] {$\frac{H}{4} \times \frac{W}{4} \times \frac{D}{4} \times 32$} (c1.west);

        \node [draw, thick, fill=white, rounded corners, below=1.5cm of block4, anchor=west, xshift=-1cm] (legend) {
            \begin{tabular}{ll ll}
                \tikz\draw[subbox, patch_color] (0,0) rectangle (0.4,0.2);      & Patch Embedding &
                \tikz\draw[subbox, epa_color] (0,0) rectangle (0.4,0.2);        & EPA Block             \\
                \tikz\draw[subbox, down_color] (0,0) rectangle (0.4,0.2);       & Downsampling    &
                \tikz\draw[subbox, up_color] (0,0) rectangle (0.4,0.2);         & Upsampling            \\
                \tikz\draw[subbox, conv_color] (0,0) rectangle (0.4,0.2);       & ConvBlock       &
                \tikz\draw[subbox, final_conv_color] (0,0) rectangle (0.4,0.2); & 1x1x1 Conv            \\
                \tikz\draw[concat] (0,0) circle (0.15) node {};                 & Concatenation   &   &
            \end{tabular}
        };

    \end{tikzpicture}
}
	\caption{Schematic visualization of the UNETR++ architecture as used in this benchmark. \cite{0000-06}}
	\label{Fig-5}
\end{figure}
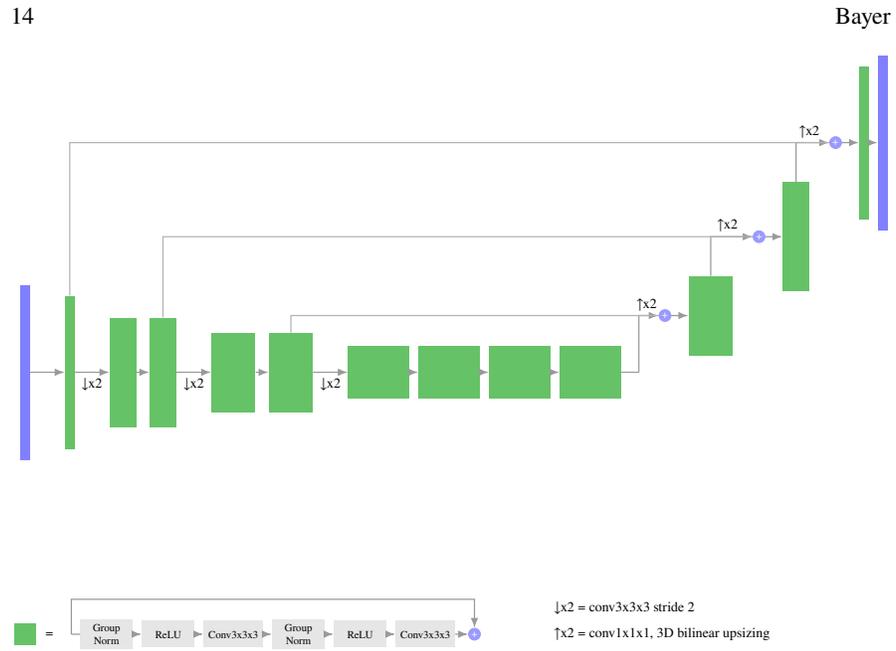
\begin{figure}[ht]
	\centering
	\resizebox{\textwidth}{!}{

    \begin{tikzpicture}[
            node distance=0.5cm,
            blue_thin/.style={rectangle, fill=blue!50, minimum width=0.15cm, minimum height=4cm},
            green_thin/.style={rectangle, fill=green!60!black!60, minimum width=0.15cm, minimum height=3.5cm},
            green_med/.style={rectangle, fill=green!60!black!60, minimum width=0.6cm, minimum height=2.5cm},
            green_wide/.style={rectangle, fill=green!60!black!60, minimum width=1.0cm, minimum height=1.8cm},
            green_widest/.style={rectangle, fill=green!60!black!60, minimum width=1.4cm, minimum height=1.2cm},
            latent_box/.style={rectangle, fill=white, minimum width=0.8cm, minimum height=0.6cm},
            plus/.style={circle, draw=blue!40, fill=blue!40, inner sep=1pt, text=white, font=\boldmath, scale=0.8},
            arrow/.style={-Latex, thick, gray!80},
            resnet_unit/.style={rectangle, fill=gray!20, minimum width=1.2cm, minimum height=0.6cm, font=\scriptsize, align=center},
        ]

        \node[blue_thin] (in) at (0,0) {};

        \node[green_thin, right=0.8cm of in] (e1) {};
        \draw[arrow] (in) -- (e1);

        \node[green_med, right=0.8cm of e1] (e2) {};
        \draw[arrow] (e1) -- (e2) node[midway, below, black, font=\small] {$\downarrow$x2};
        \node[green_med, right=0.3cm of e2] (e3) {};
        \draw[arrow] (e2) -- (e3);

        \node[green_wide, right=0.8cm of e3] (e4) {};
        \draw[arrow] (e3) -- (e4) node[midway, below, black, font=\small] {$\downarrow$x2};
        \node[green_wide, right=0.3cm of e4] (e5) {};
        \draw[arrow] (e4) -- (e5) ;

        \node[green_widest, right=0.8cm of e5] (b1) {};
        \draw[arrow] (e5) -- (b1) node[midway, below, black, font=\small] {$\downarrow$x2};
        \node[green_widest, right=0.2cm of b1] (b2) {};
        \node[green_widest, right=0.2cm of b2] (b3) {};
        \node[green_widest, right=0.2cm of b3] (b4) {};
        \draw[arrow] (b1) -- (b2); \draw[arrow] (b2) -- (b3); \draw[arrow] (b3) -- (b4);

        \coordinate (split) at ($(b4.east)+(0.4,0)$);
        \draw[thick, gray!80] (b4.east) -- (split);

        \node[plus, above right=1.2cm and 0.5cm of split] (c1) {+};
        \draw[arrow] (split) |- (c1) node[pos=0.7, above, black, font=\small] {$\uparrow$x2};

        \node[green_wide, right=0.4cm of c1] (ub1) {};
        \draw[arrow] (c1) -- (ub1);

        \node[plus, above right=0.8cm and 0.5cm of ub1] (c2) {+};
        \draw[arrow] (ub1) |- (c2) node[pos=0.7, above, black, font=\small] {$\uparrow$x2};

        \node[green_med, right=0.4cm of c2] (ub2) {};
        \draw[arrow] (c2) -- (ub2);

        \node[plus, above right=0.8cm and 0.5cm of ub2] (c3) {+};
        \draw[arrow] (ub2) |- (c3) node[pos=0.7, above, black, font=\small] {$\uparrow$x2};

        \node[green_thin, right=0.4cm of c3] (ub3) {};
        \draw[arrow] (c3) -- (ub3);
        \node[blue_thin, right=0.2cm of ub3] (final_blue) {};
        \draw[arrow] (ub3) -- (final_blue);






        \draw[gray!60, thick] (e1.north) |- (c3.west);
        \draw[gray!60, thick] (e3.north) |- (c2.west);
        \draw[gray!60, thick] (e5.north) |- (c1.west);

        \begin{scope}[shift={(0,-6)}]
            \node[rectangle, fill=green!60!black!60, minimum size=0.5cm] (green_sq) at (0,0) {};
            \node[right=0.1cm of green_sq] (eq) {=};

            \coordinate (entry) at ($(eq.east)+(0.3,0)$);
            \node[resnet_unit, right=0.2cm of entry] (gn1) {Group\\Norm};
            \node[resnet_unit, right=0.2cm of gn1] (re1) {ReLU};
            \node[resnet_unit, right=0.2cm of re1] (cv1) {Conv3x3x3};
            \node[resnet_unit, right=0.2cm of cv1] (gn2) {Group\\Norm};
            \node[resnet_unit, right=0.2cm of gn2] (re2) {ReLU};
            \node[resnet_unit, right=0.2cm of re2] (cv2) {Conv3x3x3};
            \node[plus, right=0.3cm of cv2] (add) {+};

            \draw[arrow] (gn1) -- (re1); \draw[arrow] (re1) -- (cv1);
            \draw[arrow] (cv1) -- (gn2); \draw[arrow] (gn2) -- (re2);
            \draw[arrow] (re2) -- (cv2); \draw[arrow] (cv2) -- (add);

            \draw[arrow] (entry) -- ++(0,0.8) -| (add);
            \draw[thick, gray!80] (entry) -- (gn1);
        \end{scope}

        \begin{scope}[shift={(12,-5.8)}]
            \node[anchor=west, font=\small] at (0,0.4)  {$\downarrow$x2 = conv3x3x3 stride 2};
            \node[anchor=west, font=\small] at (0,-0.2) {$\uparrow$x2 = conv1x1x1, 3D bilinear upsizing};
        \end{scope}

    \end{tikzpicture}
}
	\caption{Schematic visualization of the SegResNet architecture as used in this benchmark. \cite{0000-07}}
	\label{Fig-6}
\end{figure}

\end{document}